\RequirePackage{fix-cm}
\documentclass[twocolumn]{svjour3}          
\smartqed  
\usepackage{graphicx}
\usepackage{cite}
\usepackage{amsmath,amssymb,amsfonts}
\usepackage{algorithmic}
\usepackage{textcomp}
\usepackage{subfigure} 

\newcommand{\ie}{\textit{i.e.}\@ }
\usepackage{multirow}
\usepackage{wrapfig}
\usepackage{tikz}

\begin{document}

\title{PConv: Simple yet Effective Convolutional Layer \\for Generative Adversarial Network}


\author{Seung Park \and Yoon-Jae Yeo \and Yong-Goo Shin }

\institute{S. Park \at
              Biomedical Engineering, Chungbuk National University Hospital, 776, Seowon-gu, Cheongju-si, Chungcheongbuk-do, Rep. of Korea\\
              \email{spark.cbnuh@gmail.com}
           \and
           Y.-J. Yeo \at
              School of Electrical Engineering Department, Korea University, Anam-dong, Sungbuk-gu, Seoul, 136-713, Rep. of Korea\\
              \email{yjyeo@dali.korea.ac.kr}
           \and
           Y.-G. Shin  \at
              Division of Smart Interdisciplinary Engineering, Hannam University, Daedeok-Gu, Daejeon, 34430, Rep. of Korea\\
              \email{ygshin@hnu.kr (corresponding author)}
}

\date{Received: date / Accepted: date}

\maketitle
\begin{abstract}
This paper presents a novel convolutional layer, called perturbed convolution (PConv), which focuses on achieving two goals simultaneously: improving the generative adversarial network (GAN) performance and alleviating the memorization problem in which the discriminator memorizes all images from a given dataset as training progresses. In PConv, perturbed features are generated by randomly disturbing an input tensor before performing the convolution operation. This approach is simple but surprisingly effective. First, to produce a similar output even with the perturbed tensor, each layer in the discriminator should learn robust features having a small local Lipschitz value. Second, since the input tensor is randomly perturbed during the training procedure like the dropout in neural networks, the memorization problem could be alleviated. To show the generalization ability of the proposed method, we conducted extensive experiments with various loss functions and datasets including CIFAR-10, CelebA, CelebA-HQ, LSUN, and tiny-ImageNet. The quantitative evaluations demonstrate that PConv effectively boosts the performance of GAN and conditional GAN in terms of Frechet inception distance (FID). 

\end{abstract}

\section{Introduction}
\label{sec1}
Generative adversarial network (GAN)~\cite{goodfellow2014generative}, which is based on convolutional neural networks (CNNs), have achieved rapid advancements in various applications such as image inpainting~\cite{yu2018free, shin2020pepsi++, sagong2019pepsi}, image-to-image translation~\cite{isola2017image, choi2018stargan, zhu2017unpaired}, and text-to-image translation~\cite{reed2016generative, hong2018inferring}. However, this great success still suffers from one major problem: instability in the training procedure~\cite{salimans2016improved}. Since a goal of GAN training is to find the Nash equilibrium of a non-convex game in a continuous and high dimensional parameter space, GAN is substantially more complicated and difficult to train, compared to neural networks that are based on supervised learning~\cite{zhang2019consistency}. To alleviate this problem, some researchers~\cite{karras2017progressive, zhang2018self, zhang2018stackgan++, brock2018large} propose novel network architectures for discriminator and generator. Although these methods successfully generate high-resolution images on challenging datasets such as ImageNet~\cite{krizhevsky2012imagenet}, they still have the fundamental problem of the instability of GAN training.

Instead of modifying the network architecture, various studies~\cite{zhang2019consistency, gulrajani2017improved, miyato2018spectral, roth2017stabilizing, mescheder2018training, zhou2018don, kodali2017convergence} proposed normalization and regularization techniques that penalize the discriminator for alleviating the instability of GAN training. The most widely used normalization technique is spectral normalization~\cite{miyato2018spectral}, which imposes the Lipschitz constraint by dividing weight matrices of the discriminator with an approximation of their largest singular value. As a regularization, Gulrajani~\textit{et al.}~\cite{gulrajani2017improved} introduced the gradient regularization, called gradient penalty, which penalizes the gradient norm of straight lines between real and generated samples. Kodali~\textit{et al.}~\cite{kodali2017convergence} proposed another form of gradient regularization which constrains the magnitude of the gradient as one around the real samples. Roth~\textit{et al.}~\cite{roth2017stabilizing} presented a stabilizing regularization technique that directly regularizes the squared gradient norm, where the gradient is calculated with respect to the real and generated samples. These normalization and regularization techniques are effective to improve the performance of GAN. However, some researchers~\cite{kurach2019large, zhang2019consistency} pointed out that when both normalization and gradient-based regularization are used, the performance is either slightly improved or it fails to improve. 

Recent studies~\cite{brock2018large, zhao2020differentiable} argued that the memorization problem is another reason for the instability of the GAN training. As mentioned in~\cite{zhao2020differentiable}, when the discriminator memorizes all images from a given dataset as training progresses, \textit{i.e.} the memorization problem occurs, it disrupts the training dynamics and degrades a generated image quality. Brock~\textit{et al.}~\cite{brock2018large} observed that this severe problem has not only happened on small datasets; the memorization problem often occurs on large-scale datasets including ImageNet. To alleviate this problem, some researchers~\cite{zhao2020differentiable, karras2020training, tran2020towards, zhao2020image} applied data augmentation techniques such as translation, zoom-in/out, and Cutout~\cite{devries2017improved}. These approaches effectively prevent the memorization problem and improve GAN performance. 

Despite the extensive ongoing efforts to develop the normalization, regularization, and data augmentation techniques, there are still some fundamental challenges. To the best of our knowledge, there are no previous works that attempt to develop a convolutional layer for alleviating these problems. In this paper, we propose the new form of the convolutional layer specialized for discriminator, called perturbed convolution (PConv), which aims at achieving two goals simultaneously: boosting the generative adversarial network (GAN) performance and moderating the memorization problem where the discriminator memorizes all images from a given dataset as the training progresses. The proposed method produces perturbed features by randomly disturbing an input tensor prior to performing the convolution operation. PConv is simple but surprisingly effective. First, to make similar output even with the perturbed tensor, each layer in the discriminator should learn robust features having a small local Lipschitz value. Second, when the input tensor is randomly perturbed during the training procedure like the dropout in neural networks, the memorization problem can be alleviated. By replacing the standard convolutional layer with the perturbed-convolutional layer, the proposed method can be easily applied to existing network architectures without imposing training overheads or additional computational cost. To demonstrate the generalization ability of the proposed method, we conducted series of experiments with various datasets including CIFAR-10, CelebA, CelebA-HQ, LSUN, and tiny-Image Net. The quantitative evaluations show that the proposed method significantly improves the performance of GAN and conditional GAN in terms of Frechet inception distance (FID). 

In summary, our contributions of the study are summarized as follows. First, we propose a novel convolutional layer, \ie PConv, which can be easily applied to the existing GAN without modifying the network architectures. Second, the proposed method significantly boosts the performance of GAN without training overhead or additional computational cost. Third, we conducted extensive ablation studies to demonstrate the generalization ability of the proposed method. In various datasets and experimental settings, GAN with the proposed method shows a superior performance than GAN with the standard convolutional layer. 

\begin{figure*}
\centering
\includegraphics[width=0.85\linewidth]{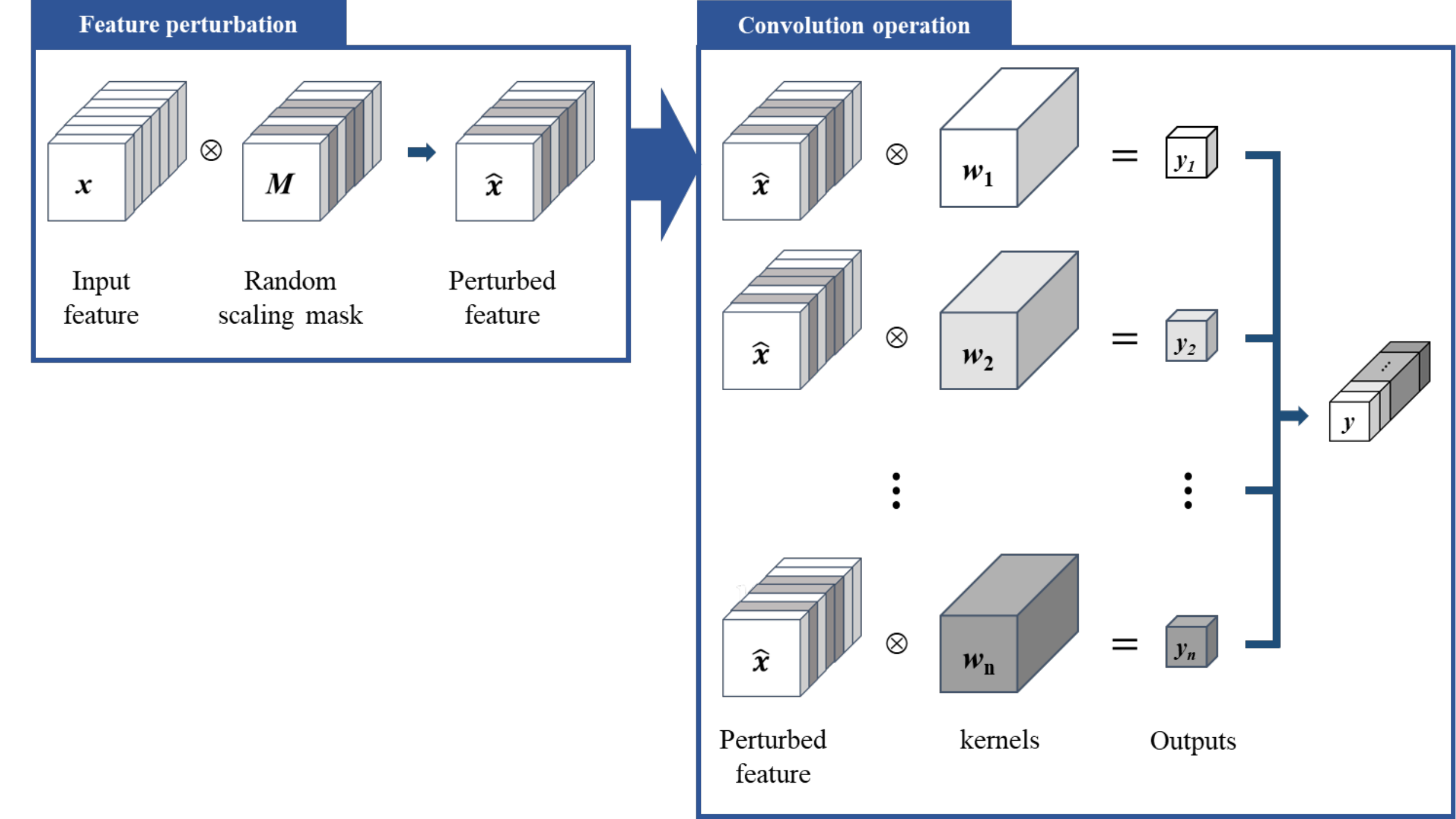}
\caption{Overall framework of the proposed method. Contrary to the standard convolutional layer, the proposed method disturbs the input tensor before conducting the convolutional operation to produce the perturbed features.}
\label{fig:fig1}
\vspace{-0cm}
\end{figure*}

\section{Background}
\label{sec2}
\subsection{Generative Adversarial Network}
\label{sec2.1}
In general, GAN~\cite{goodfellow2014generative} consists of generator $G$ and discriminator $D$. In the original setting, both networks are trained simultaneously; nonetheless, their goals are different. $G$ is optimized to produce visually appealing samples, whereas $D$ is trained to distinguish the generated samples from real ones. This procedure can be summarized as the following objective functions:

\begin{eqnarray}
\label{eq1:ganD}
       \lefteqn{L_D = - E_{x\sim P_\textrm{data}(x)}[\log D(x)]}\nonumber\\
    & & {\qquad \qquad \qquad} - E_{z\sim P_{z(z)}}[\log(1-D(G(z)))],
\end{eqnarray}

\begin{eqnarray}
\label{eq1:ganG}
    L_G = -E_{z\sim P_{z(z)}}[\log(D(G(z)))],
\end{eqnarray}
where $L_D$ and $L_G$ are the objective function for the discriminator and generator, respectively. In addition, $z$ and $x$ indicate a random noise vector and a real sample from the random normal distribution $P_z(z)$ and the data distribution $P_{data}(x)$, respectively. To improve the stability of the training process, several studies on the modification of the equations~\ref{eq1:ganD} and~\ref{eq1:ganG} have benn conducted. For instance, Mao~\textit{et al.}~\cite{mao2017least} applied the least square errors to the objective function (LSGAN), whereas Arjovsky~\textit{et al.}~\cite{arjovsky2017wasserstein} computed the loss value by measuring the Wasserstein distance between the real and generated distributions (WGAN). Another commonly adopted GAN formulation is the hinge-version of adversarial loss~\cite{lim2017geometric}, which is written as

\begin{eqnarray}
\label{eq2:hinge}
    \lefteqn{L_D=E_{x\sim P_{data}(x)}[\max(0, 1-D(x))]} \nonumber\\
    & & {\qquad \qquad \qquad} + E_{z\sim P_{z(z)}}[\max(0, 1+D(G(z)))],
\end{eqnarray}

\begin{equation}
    L_G=-E_{z\sim P_{z(z)}}[D(G(z))].
\label{eq3:hinge}
\end{equation}
The current widely-used practice is to employ the hinge-version of adversarial loss while enforcing spectral normalization~\cite{miyato2018spectral} either only on the discriminator or on both the discriminator and generator. 

On the other hand, conditional GAN which focuses on producing the class conditional images has been actively researched~\cite{mirza2014conditional, odena2017conditional, miyato2018spectral, zhang2017stackgan}. The conditional GAN usually employs conditional information $c$, \textit{e.g.} class labels or text condition, to both generator and discriminator in order to control the data generation process. This procedure can be formulated as follows:

\begin{eqnarray}
\label{eq2:cganD}
       \lefteqn{L_D = - E_{(x,c)\sim P_\textrm{data}(x)}[\log D(x, c)]}\nonumber\\
    & & {\qquad \quad} - E_{z\sim P_{z(z)},c\sim P_{\mathrm{data}}(x)}[\log(1-D(G(z, c)))],
\end{eqnarray}

\begin{eqnarray}
\label{eq2:cganG}
    L_G = - E_{z\sim P_{z(z)},c\sim P_{\mathrm{data}}(x)}[\log(D(G(z, c)))],
\end{eqnarray}
By training the networks based on Eqs.~\ref{eq2:cganD} and~\ref{eq2:cganG}, the generator can select an image category to be generated, which is not possible when employing the GAN framework.


\subsection{Memorization problem in the discriminator}
\label{sec2.3}
To avoid the overfitting problem, deep learning models in various fields usually adopt label-preserving data augmentation techniques such as region masking~\cite{devries2017improved}, flipping, rotation, cropping~\cite{krizhevsky2017imagenet, wan2013regularization}, data mixing~\cite{zhang2017mixup}, and local and affine distortions~\cite{simard2003best}. Inspired by these approaches, to moderate the memorization problem in the discriminator, some researchers~\cite{zhao2020differentiable, zhang2019consistency, karras2020training, tran2020towards, zhao2020image} apply data augmentation techniques for training GAN. Particularly, Zhao~\textit{et el.}~\cite{zhao2020differentiable} conducted extensive experiments with various cases and demonstrated that data augmentation techniques is effective to avoid the memorization problem in the discriminator. 

On the other hand, instead of using the data augmentation methods, Brock~\textit{et al.}~\cite{brock2018large} applied the dropout technique~\cite{srivastava2014dropout} to the last layer in the discriminator for alleviating the memorization problem. However, they argued that the traditional dropout strategy could alleviate the memorization problem; however it degrades the performance of the GAN. In this paper, we also attempted to apply the conventional dropout-based approach for moderating the memorization problem, but we observed that the performance was drastically degraded when the dropout ratio was increased. These observations indicate that the conventional dropout techniques are not suitable to train the GAN.  

\begin{figure*}
\centering
\includegraphics[width=0.9\linewidth]{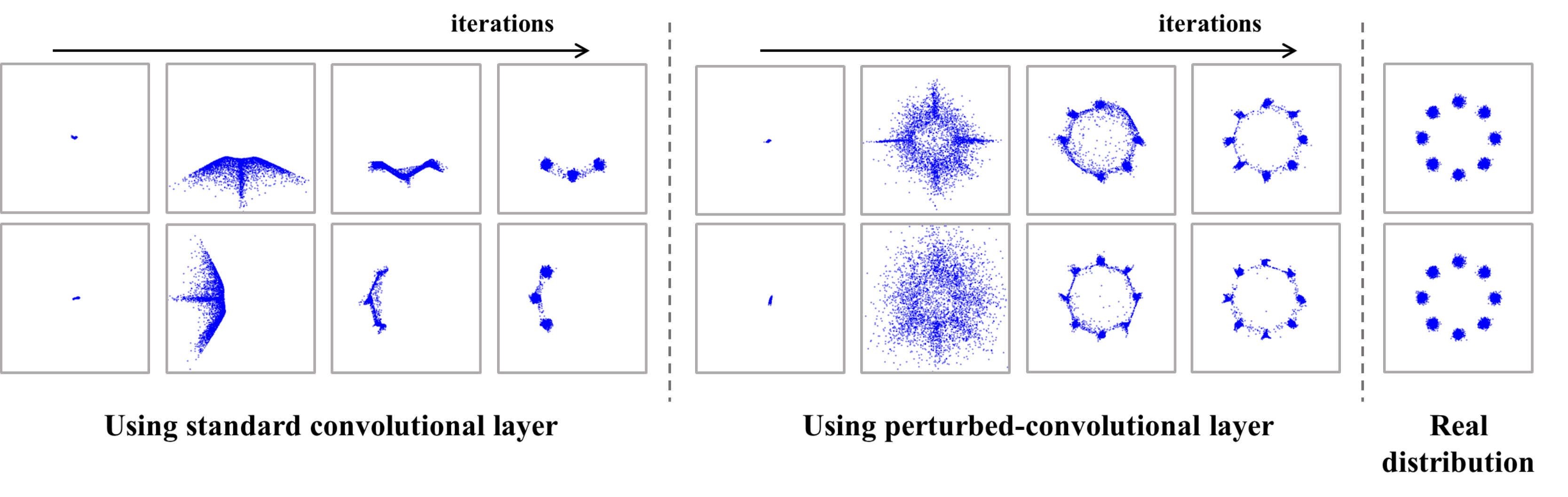}
\caption{Illustration of the GAN training results on eight 2D Gaussian mixture models. We have trained the networks two times to reveal the trend of the perturbed-convolutional layer module.}
\label{fig:fig2}
\vspace{-0cm}
\end{figure*}

\section{Proposed Method}
\label{sec3}
In this paper, we propose a novel convolutional layer that not only moderates the memorization problem similar to dropout techniques but also effectively improves the performance of GAN. Fig.~\ref{fig:fig1} shows the overall framework of the proposed method. As depicted in Fig.~\ref{fig:fig1}, before conducting the convolutional operation, the proposed method randomly disturbs the input tensor by multiplying a random scaling mask \textit{M} in which randomly selected channels have a random constant value $k \in [0, 1]$, others have one. In particular, \textit{M} is built as follows: a tensor consisting of the constant value 1 is first produced. Then, such as the channel-selecting procedure in the dropout technique, some channels to be perturbed are randomly selected by a certain ratio $\lambda_\textrm{R}$. After that the selected channels are scaled down by multiplying the common \textit{k}. By performing these simple procedures, we could build \textit{M} that has $k$ in the randomly selected channels, others have the value 1. 

Therefore, PConv with the input feature \textit{x} can be defined as follows:

\begin{eqnarray}
\label{eq4}
    y_i = f_{i}(x) = (x\otimes M)*w_i,
\end{eqnarray}
where $f_{i}(\cdot)$ and $w_i$ indicate the \textit{i-th} output value and convolutional kernel, respectively. Tensor broadcasting is included in Eq.~\ref{eq4} and different random scaling masks are employed to each intermediate layers. This masking operation is simple and meaningful for the discriminator. To classify real and generated images even if the input tensor is randomly perturbed, PConv should learn the robust features that do not have much effect on the subsequent convolutional layer. In other words, to minimize the adversarial loss successfully, $|f_{i}(x)-f_{i}(\hat x)|$ should become a small value, where $\hat x$ indicates the perturbed feature, \ie $x\otimes M$. This indicates that $f_{i}(\cdot)$ should have a small local Lipschitz constant; if the Lipschitz constant of $f_{i}( \cdot)$ is large, the output of the discriminator will change a lot with small perturbations. Therefore, by simply replacing the standard convolutional layer with PConv, it is possible to lead the discriminator to learn robust features having a small local Lipschitz constant. Note that PConv is not designed to constraint the global Lipschitz constant of the discriminator; it is designed to learn robust features in each layer. 

\begin{figure}
\centering
\includegraphics[width=0.95\linewidth]{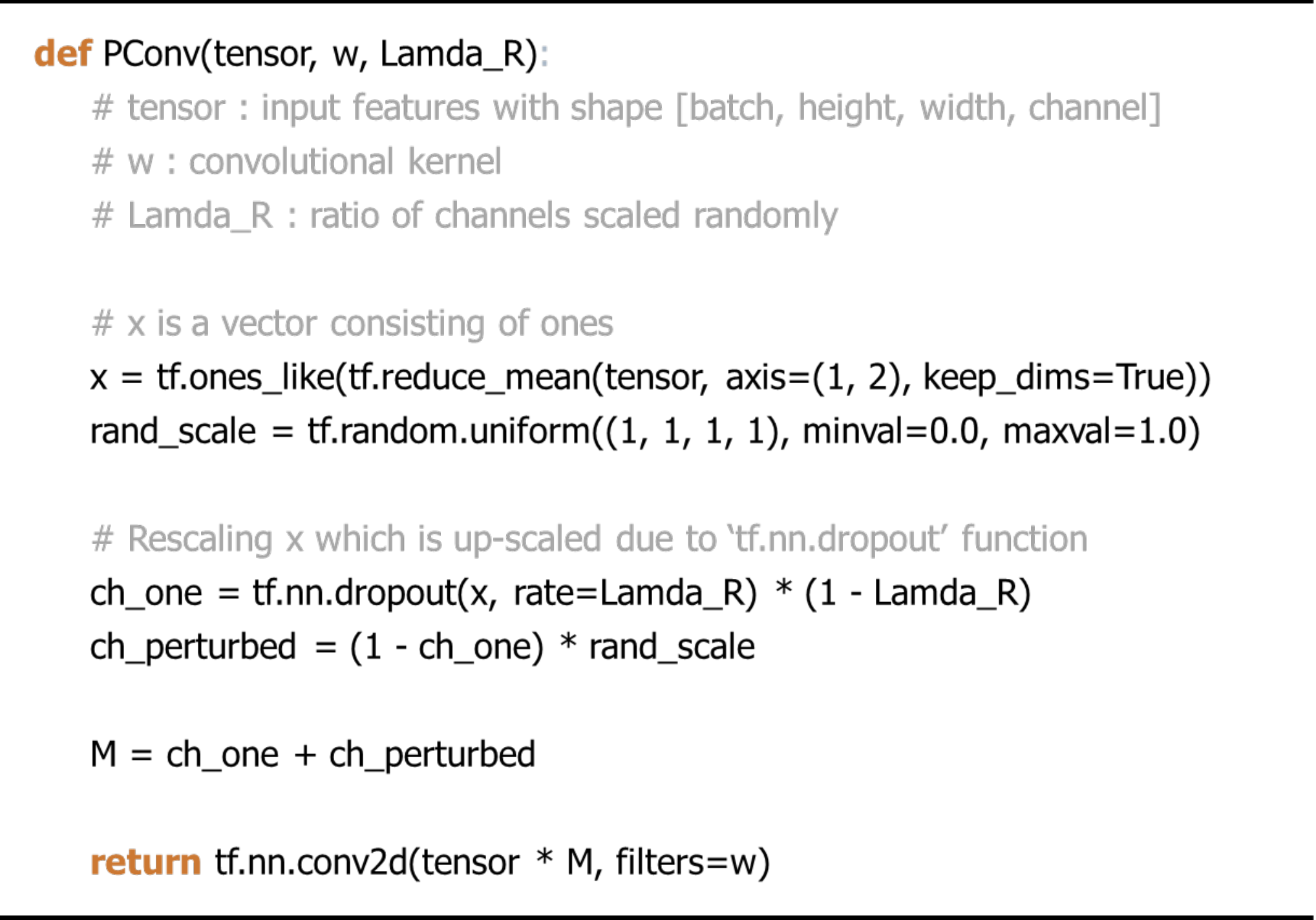}
\caption{Python code of perturbed-convolutional layer based on \textit{TensorFlow}.}
\label{fig:fig3}
\vspace{-0cm}
\end{figure}

To validate the effectiveness of the proposed method, we trained a GAN with a simple network architecture consisting of multiple fully-connected layers on eight two-dimensional (2D) Gaussian mixture models (GMMs). For a fair comparison, in the discriminator, we only replaced the fully-connected layers with the perturbed-fully-connected layers. Fig.~\ref{fig:fig2} illustrates the experimental results. As the training process continues, the GAN consisting of standard fully-connected layers suffers from the mode collapse problem, whereas the GAN with the perturbed-fully-connected layers learns all the GMMs successfully. These results indicate that the proposed method is superior to the standard convolutional layer in GAN training. More extensive experiments will be presented in the next section. The detailed implementation code for PConv based on \textit{Tensorflow} is described in Fig.~\ref{fig:fig3}. We only need to build the random scaling mask before performing the convolutional layer. This means since PConv only contains a trivial multiplication operation that can be computed quickly, the proposed method does not incur training overhead; the multiplication operation is slight compared to the convolution operation, thus we could argue that PConv does not impose the training overhead. In addition, since PConv is used for the discriminator, it does not affect the test phase that generates images from the generator.

Indeed, one may anticipate that the PConv is similar to the conventional spatial-dropout (SDrop) method \cite{tompson2015efficient}, which drops out the randomly selected channels of the input feature. However, there is a major difference between the proposed method and the conventional one: the existence of the random scaling value \textit{k}. More specifically, SDrop sets the randomly selected channels to zero, whereas the proposed method scales down the features in those channels by multiplying \textit{k}. This small difference has a large effect on GAN training. To prove the theoretical validity of this assumption, we will show an example. Let us consider \textbf{x} as an \textit{n}-dimensional vector, \ie $\textbf{\textrm{x}}=\{x_1, ..., x_n\}^\textrm{T}$ where $x_1\geq ...\geq x_n$, and the output of the convolutional layer, \ie \textit{y}, is a single scalar value. This indicates that the convolutional layer contains a single kernel vector $\textbf{\textrm{w}}=\{w_1, w_2, ..., w_n\}^\textrm{T}$. When the SDrop method is applied to this convolutional layer, the variation range of \textit{y}, $\Delta y$, can be defined as follows:

\begin{eqnarray}
\label{eq5}
    \textrm{\textbf{d}}_{\textrm{min}}^\textrm{T}\textbf{\textrm{w}} \leq \Delta y \leq \textrm{\textbf{d}}_{\textrm{max}}^\textrm{T}\textbf{\textrm{w}},
\end{eqnarray}
where $\textrm{\textbf{d}}_{\textrm{min}}$ and $\textrm{\textbf{d}}_{\textrm{max}}$ are the perturbed vectors which produce the minimum and maximum changes when they are projected onto \textbf{w}, respectively. For instance, when \textit{n} is 10 and $\lambda_\textrm{R}$ is set to 0.1, $\textrm{\textbf{d}}_{\textrm{min}} = \{x_1, ..., x_9, 0\}$ and $\textrm{\textbf{d}}_{\textrm{max}} = \{0, x_2, ..., x_{10}\}$. In contrast, when applying PConv, the range of $\Delta y$ becomes

\begin{eqnarray}
\label{eq6}
    0 \leq \Delta y \leq \textrm{\textbf{d}}_{\textrm{max}}^\textrm{T}\textbf{\textrm{w}},
\end{eqnarray}
because $\textrm{\textbf{d}}_{\textrm{min}}$ is zero when the random scaling value is one. As described in Eqs.~\ref{eq5} and~\ref{eq6}, the  PConv can produce the output without variations, but SDrop is not. In other words, since there is at least $\textrm{\textbf{d}}_{\textrm{min}}^\textrm{T}\textbf{\textrm{w}}$ variant in the features, the discriminator using SDrop is difficult to produce the decision boundaries that guide the generator well. In addition, when the dropout ratio becomes large, the minimum boundary of $\Delta y$ is increased in the case of SDrop, but PConv does not change. Therefore, the proposed method is less sensitive to the dropout ratio, compared to SDrop.

\begin{figure}
\centering
\includegraphics[width=0.95\linewidth]{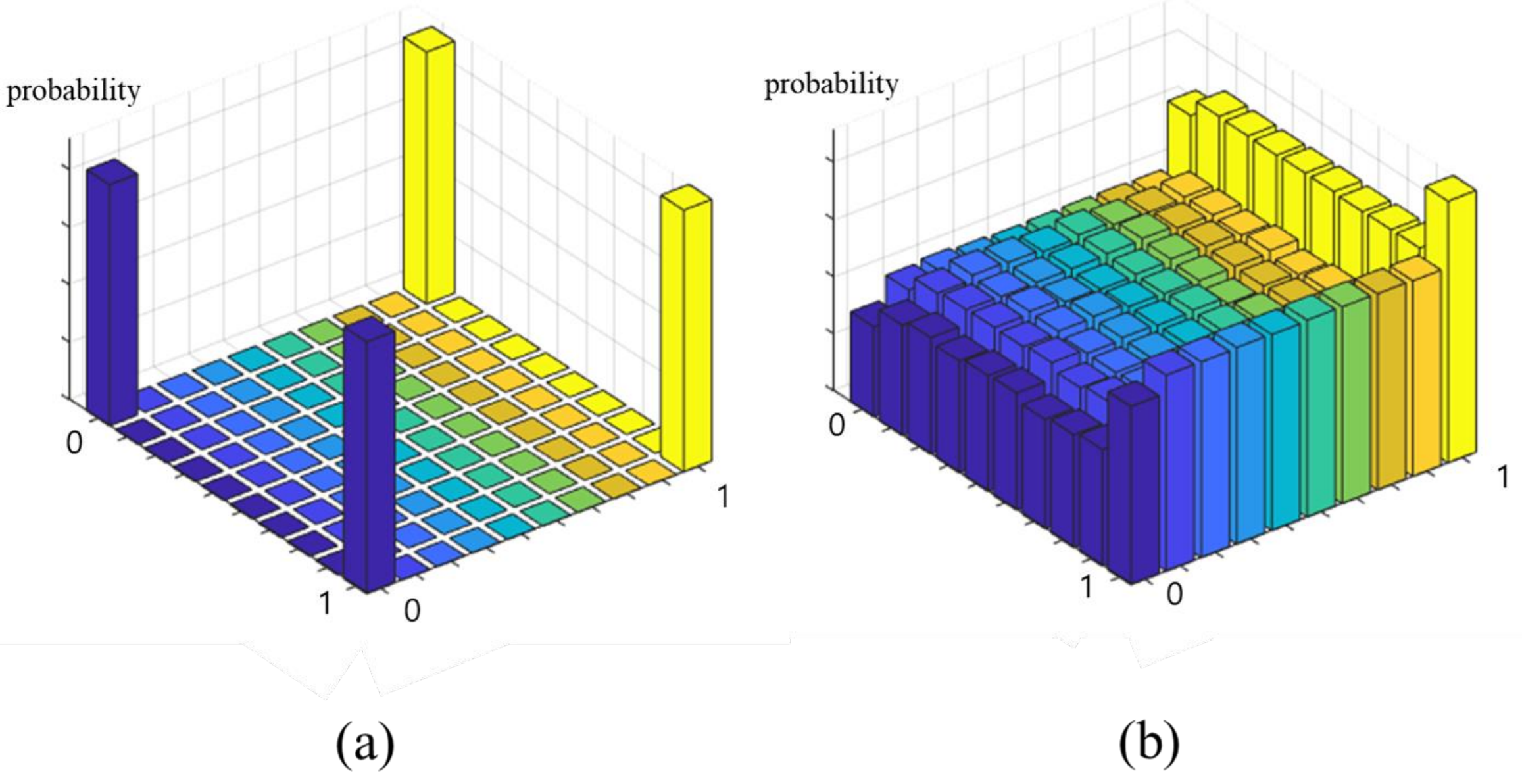}
\caption{Probability maps that represent the feature spaces when applying SDrop~\cite{tompson2015efficient} and PConv. (a) SDrop~\cite{tompson2015efficient} (b) PConv.}
\label{fig:probmap}
\vspace{-0cm}
\end{figure}

To reveal the effectiveness of the random scaling operation in PConv, we conducted toy examples that compare the output vectors of PConv and those of SDrop. In our experiments, we set the input feature $\textbf{\textrm{v}} \in \mathbb{R}^2 $ as $\{1, 1\}$ and scaled-down the randomly selected channels. More specifically, we multiplied zero or the random constant value $k \in [0, 1]$ when producing the output vectors of SDrop or PConv, respectively. Fig.~\ref{fig:probmap} shows the probability maps representing the feature spaces that SDrop and PConv can cover. As depicted in Fig.~\ref{fig:probmap}, SDrop generates discrete vectors, \ie $\{0, 0\}, \{0, 1\}, \{1, 0\}$, which are confined to the corners of the plane, whereas PConv produces continuous vectors that handle the entire plane. Although $\textrm{\textbf{d}}_{\textrm{max}}^\textrm{T}\textbf{\textrm{w}}$ values of SDrop and PConv are the same, the covered feature space is different. Since the output vectors of PConv are perturbed smoothly, it is more effective to guide the generator than the SDrop.

\begin{table*}[t]
\caption{Network architecture of the generator for each image resolution.}
\begin{center}
\begin{tabular}{c | c | c | c | c}
\hline\hline
$32\times32$ resolution & $64\times64$ resolution & $128\times128$ resolution & $256\times256$ resolution & $512\times512$ resolution \\
\hline
$z \in \mathbb{R}^{128} \sim N(0, I)$ & $z \in \mathbb{R}^{128} \sim N(0, I)$ & $z \in \mathbb{R}^{128} \sim N(0, I)$ & $z \in \mathbb{R}^{128} \sim N(0, I)$ & $z \in \mathbb{R}^{128} \sim N(0, I)$\\
FC, $4 \times 4 \times 256$ & FC, $4 \times 4 \times 512$ & FC, $4 \times 4 \times 512$ & FC, $4 \times 4 \times 512$ & FC, $4 \times 4 \times 512$ \\
ResBlock, up, 256 & ResBlock, up, 512 & ResBlock, up, 512 & ResBlock, up, 512 & ResBlock, up, 512\\
ResBlock, up, 256 & ResBlock, up, 256 & ResBlock, up, 512 & ResBlock, up, 512 & ResBlock, up, 512\\
ResBlock, up, 256 & ResBlock, up, 128 & ResBlock, up, 256 & ResBlock, up, 256 & ResBlock, up, 256\\
BN, ReLU & ResBlock, up, 64 & ResBlock, up, 128 & ResBlock, up, 128 & ResBlock, up, 128\\
$3\times3$ conv, Tanh & BN, ReLU & ResBlock, up, 64 & ResBlock, up, 64 & ResBlock, up, 64\\
 & $3\times3$ conv, Tanh & BN, ReLU & ResBlock, up, 32 & ResBlock, up, 32\\
& & $3\times3$ conv, Tanh & BN, ReLU  & ResBlock, up, 16\\
& & & $3\times3$ conv, Tanh  & BN, ReLU \\
& & & & $3\times3$ conv, Tanh   \\

\hline\hline
\end{tabular}
\end{center}
\label{table:table1}
\end{table*}

\begin{table*}[t]
\caption{Network architecture of the discriminator for each image resolution.}
\begin{center}
\begin{tabular}{c | c | c | c | c}
\hline\hline
$32\times32$ resolution & $64\times64$ resolution & $128\times128$ resolution & $256\times256$ resolution & $512\times512$ resolution \\
\hline
RGB image & RGB image & RGB image & RGB image & RGB image \\
ResBlock, down, 128 & ResBlock, down, 64 & ResBlock, down, 64 & ResBlock, down, 32 & ResBlock, down, 16\\
ResBlock, down, 128 & ResBlock, down, 128 & ResBlock, down, 128 & ResBlock, down, 64 & ResBlock, down, 32\\
ResBlock, 128 & ResBlock, down, 256 & ResBlock, down, 256 & ResBlock, down, 128 & ResBlock, down, 64\\
ResBlock, 128 & ResBlock, down, 512 & ResBlock, down, 512 & ResBlock, down, 256 & ResBlock, down, 128\\
ReLU & ResBlock, 512 & ResBlock, down, 512 & ResBlock, down, 512 & ResBlock, down, 256\\
Global sum pooling & ReLU & ResBlock, 512 & ResBlock, down, 512 & ResBlock, down, 512\\
Dense, 1 & Global sum pooling & ReLU & ResBlock, 512 & ResBlock, down, 512\\
 & Dense, 1 & Global sum pooling & ReLU & ResBlock, 512\\
 & & Dense, 1 & Global sum pooling & ReLU\\
 & &  & Dense, 1 & Global sum pooling\\
 & &  & & Dense, 1 \\

\hline\hline
\end{tabular}
\end{center}
\label{table:table2}
\end{table*}

\section{Experiments}
\label{sec4}
\subsection{Implementation details}
\label{subsec:4.1}
To show the generalization ability of PConv, we conducted extensive experiments using various datasets including CIFAR-10~\cite{27torralba200880}, LSUN~\cite{yu15lsun}, CelebA~\cite{liu2015deep}, CelebA-HQ~\cite{liu2015deep, karras2017progressive}, and tiny-ImageNet~\cite{23deng2009imagenet, yao2015tiny} (a subset of ImageNet~\cite{23deng2009imagenet}), consisting of the 200 selected classes. Among a large number of images in LSUN, we randomly selected 30,000 images per each class for training. This indicates that in this study, we built the LSUN dataset using 300,000 images. The images in the CelebA and LSUN datasets are resized to $64\times 64$ pixels, whereas the images in the tiny-ImageNet are resized to $128\times 128$ pixels. To evaluate the generation performance of high-resolution images, we utilized CelebA-HQ by resizing the images to $256\times256$ and $512\times512$ pixels. We employed the \textit{hinge}-version loss in Eqs.~\ref{eq2:hinge} and~\ref{eq3:hinge} as the adversarial objective function. 

\begin{figure}
\centering
\includegraphics[width=0.8\linewidth]{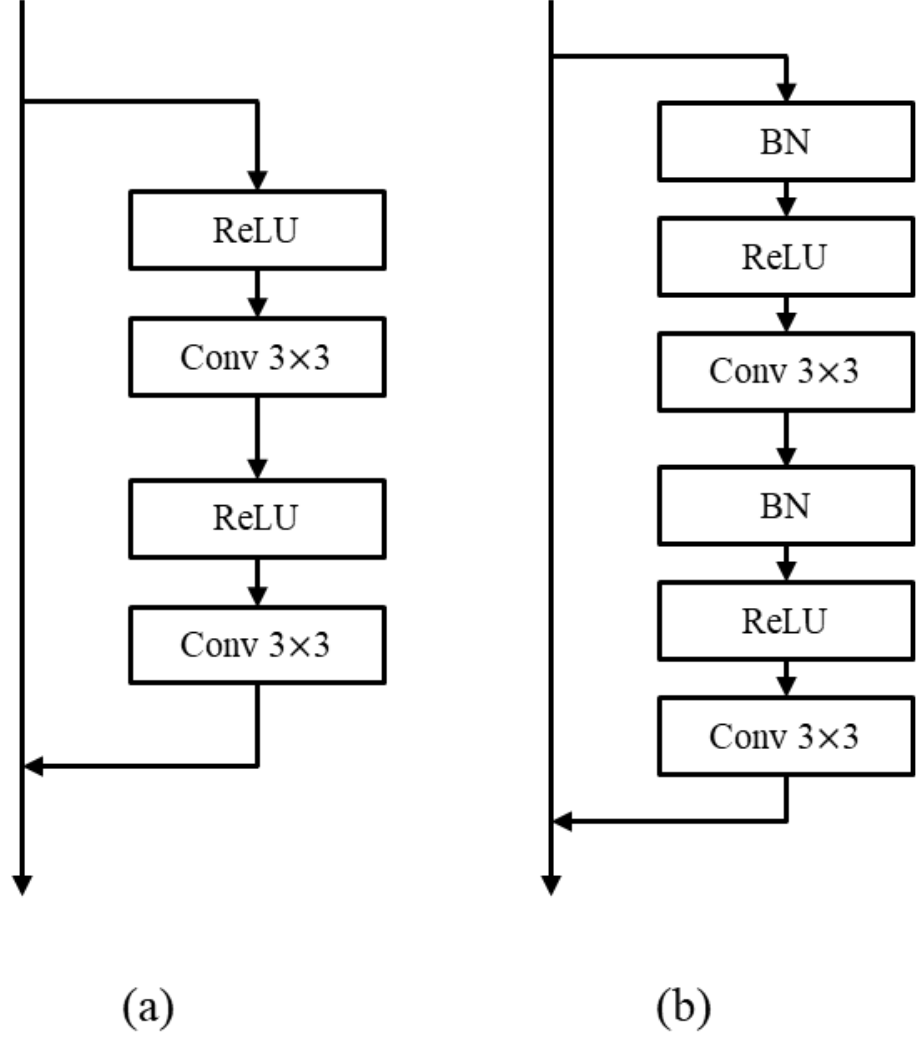}
\caption{Network architectures of ResBlock utilized in our experiments: (a) ResBlock for the discriminator and (b) ResBlock for the generator.}
\label{fig:fig4}
\vspace{-0cm}
\end{figure}

\begin{table*}[t]
\caption{Comparison of the proposed method and SDrop on CIFAR-10 in terms of FID.}
\begin{center}
\begin{tabular}{c | c | c | c | c}
\hline\hline
Method & & $\lambda_\textrm {R} = 0.1$& $\lambda_\textrm {R} = 0.2$ & $\lambda_\textrm {R} = 0.3$ \\
\hline
\multirow{4}*{SDrop} & trial 1& 12.73 & 16.96 & 88.21 \\
& trial 2 & 13.31 & 16.47 & 85.96 \\
& trial 3 & 12.99 & 16.20 & 85.30\\
\cline{2-5}
& \textbf{Average} & $13.01 \pm 0.29$ & $16.55 \pm 0.38$ & $85.30 \pm 3.29$\\

\hline
\multirow{4}*{SDrop*} & trial 1& 13.07 & 16.13 & 94.79 \\
& trial 2 & 12.65 & 16.14 & 73.56 \\
& trial 3 & 14.53 & 16.04 & 85.86 \\
\cline{2-5}
& \textbf{Average} & 13.42 $\pm$ 0.99 & 16.10 $\pm$ 0.05 & 84.74 $\pm$ 10.66\\

\hline
\multirow{4}*{$\textrm{SDrop}^\dagger$} & trial 1& 13.19 & 13.31 & 14.28 \\
& trial 2 & 13.15 & 13.45 & 13.83 \\
& trial 3 & 13.43 & 13.54 & 14.22 \\
\cline{2-5}
& \textbf{Average} & 13.26 $\pm$ 0.15 & 13.45 $\pm$ 0.08 &  \textbf{14.11 $\pm$ 0.24} \\

\hline
\multirow{4}*{PConv} & trial 1& 12.31 & 12.62 & 14.31 \\
& trial 2 & 12.58 & 12.77 & 14.43\\
& trial 3 & 12.89 & 12.45 & 13.90 \\
\cline{2-5}
& \textbf{Average} & \textbf{12.59 $\pm$ 0.29} & \textbf{12.61 $\pm$ 0.16} & 14.21 $\pm$ 0.28\\


\hline\hline
\end{tabular}
\end{center}
\label{table:table3}
\end{table*}

Since all the parameters in the generator and discriminator including (PConv) can be differentiated, we performed an optimization using the Adam optimizer \cite{kingma2014adam}, which is a stochastic optimization method with adaptive estimation of moments. We set the parameters of the Adam optimizer, \ie $\beta _1$ and $\beta _2$, to 0 and 0.9, respectively, and we set the learning rate to 0.0002. During the last 50,000 iterations of the training, we decrease the learning rate linearly. Similar to the conventional methods~\cite{miyato2018cgans, miyato2018spectral, gulrajani2017improved, arjovsky2017wasserstein}, each time we updates the generator, the discriminator was updated five times using different mini-batches. For the CIFAR-10, CelebA, and LSUN datasets, we set the batch size to 64 and trained the generator for 50k, 100k, and 100k iterations, respectively. In addition, for the CelebA-HQ and tiny-ImageNet, we trained the network 100k and 450k iterations with 16 and 32 batch sizes, respectively. It is worth noting that we trained the generator with a batch size twice as large as that of the discriminator. For instance, on the CIFAR-10 dataset, we trained the discriminator with a batch size of 64, whereas the generator was trained with a batch size of 128. The proposed method contains a single hyper-parameter $\lambda_\textrm{R}$ which determines how many channels are scaled randomly during the training procedure. How we determined $\lambda_\textrm{R}$ value will be described in Section~\ref{subsec:4.4}. Note that we perturbed the discriminator features when training the not only discriminator but also generator. More specifically, the generator is trained to synthesize images having features that are used to classify the real and generated images in the discriminator. Thus, if the features used to classify the images are different from the features used for training the generator, the adversarial learning becomes unstable. In other words, the features used to train the discriminator and generator should be the same.

In this paper, we employed the generator and discriminator architectures consisting of multiple residual blocks~\cite{he2016deep} as our baseline models~\cite{miyato2018cgans, miyato2018spectral, yeo2021simple}. The detailed network architectures for the generator and discriminator are presented in Tables~\ref{table:table1} and ~\ref{table:table2}, where ResBlock architectures are described in Fig.~\ref{fig:fig4}. In the discriminator, we utilized the spectral normalization~\cite{miyato2018spectral} for each layers. The discriminator down-samples the feature maps using the average-pooling after the second convolutional layer by using the average-pooling, whereas up-sampling (a nearest-neighbor interpolation) is performed before the first convolutional layer in the generator. 

\subsection{Performance evaluation metric}
\label{subsec:4.3}

\begin{table}[t]
\caption{Comparison of the standard convolution, SDrop, and PConv on CIFAR-10 in terms of classification accuracy.}
\begin{center}
\begin{tabular}{c | c | c | c | c | c | c}
\hline\hline
& \multicolumn{2}{c|}{Conv} & \multicolumn{2}{c|}{SDrop} & \multicolumn{2}{c}{PConv} \\
\hline
& train & test & train & test & train & test \\
\hline
Accuracy & 83.52 & 31.79 & 79.63 & 50.01 & 80.73 & 40.97 \\

\hline\hline
\end{tabular}
\end{center}
\label{table:memorization}
\end{table}

To evaluate how well the generator produces the image, we employed the most popular assessments called Frechet inception distance (FID)~\cite{heusel2017gans}. This metric measures the visual appearance and diversity of the generated images using the Wasserstein distance between the distributions of the real and generated images in the feature space obtained by the Inception model~\cite{szegedy2016rethinking}. The FID can be expressed as: 

\begin{equation}
\label{eq7}
    \textrm{FID}(p,q) = \| \mu_p - \mu_q \|_2^2 + \mathrm{trace}(C_p +C_q - 2(C_p C_q)^{1/2}),
\end{equation}\\
where $ \{\mu_p,C_p \}$ and $\{\mu_q,C_q \}$ are the mean and covariance of the samples with the distributions of the real and generated images, respectively. Lower FID scores mean better quality of the generated images. To measure the performance using the FID, in this paper, we generated 50,000 images for CIFAR-10, LSUN, CelebA, and tiny-ImageNet and 30,000 images for Celeb-HQ.


\begin{table*}[t]
\caption{Comparison of the FID scores in different sizes of the datasets on CIFAR-10.}
\begin{center}
\begin{tabular}{c | c | c | c | c}
\hline\hline
Dataset size & & Conv & Conv + DA~\cite{zhao2020differentiable} & PConv \\
\hline
\multirow{4}*{Full} & trial 1& 13.21 & 14.39 & 12.31 \\
& trial 2 & 13.44 & 14.01 & 12.58 \\
& trial 3 & 13.57 & 14.10 & 12.89 \\
\cline{2-5}
& \textbf{Average} & $13.41 \pm 0.18$ & $14.17 \pm 0.20$ & \textbf{12.59} $\pm$ \textbf{0.29}\\
\hline

\multirow{4}*{Half} & trial 1& 17.54 & 15.75 & 14.44 \\
& trial 2 & 17.47 & 16.84 & 13.50 \\
& trial 3 & 18.21 & 15.55 & 14.07 \\
\cline{2-5}
& \textbf{Average} & 17.74 $\pm$ 0.41 & 16.05 $\pm$ 0.70 & \textbf{14.00 $\pm$ 0.47} \\

\hline
\multirow{4}*{Quarter} & trial 1& 25.79 & 19.28 & 17.84 \\
& trial 2 & 27.25 & 19.86 & 19.45 \\
& trial 3 & 24.74 & 18.18 & 17.89 \\
\cline{2-5}
& \textbf{Average} & $ 25.93 \pm 1.26$ & 19.11 $\pm$ 0.85 & \textbf{18.39} $\pm$ \textbf{0.92}\\

\hline\hline
\end{tabular}
\end{center}
\label{table:table4}
\end{table*}

\subsection{Quantitative comparison}
\label{subsec:4.4}
Before evaluating the performance of PConv on the various datasets, we first present the ablation studies on the CIFAR-10 dataset. We trained the network three times from the scratch to show that the performance gain was not due to the lucky weight initialization. First, we discuss the difference between the proposed method and SDrop~\cite{tompson2015efficient}. As shown in Table~\ref{table:table3}, SDrop shows poor performance compared to the proposed method. In particular, the proposed method shows a stable performance, even with an increasing value of $\lambda_\textrm {R}$. In contrast, the performance of SDrop is drastically degraded when $\lambda_\textrm{R}$ value becomes large. To make our results more reliable, we conducted additional experiments that randomly changed the $\lambda_\textrm {R}$ of SDrop during the training procedure. This indicates that we verified whether SDrop shows fine performance when $\textrm{\textbf{d}}_{\textrm{min}}$ in Eq.~\ref{eq5} is zero. In order to statistically match the dropout ratio with other experiments, we randomly changed $\lambda_\textrm {R}$ in the range [$\lambda_\textrm {R} - 0.1$, $\lambda_\textrm {R} + 0.1$]. For instance, when $\lambda_\textrm {R}$ is set to 0.1, the statistical dropout ratio is 0.1, whereas $\textrm{\textbf{d}}_{\textrm{min}}$ becomes zero. As described in Table~\ref{table:table3}, SDrop with random dropout ratio (we denoted it as SDrop*), shows a similar trend with SDrop. Although $\textrm{\textbf{d}}_{\textrm{min}}$ becomes zero, SDrop* still converts the randomly selected channels as zero, which disturbs the adversarial learning.

\begin{table}[t]
\caption{Comparison of FID scores in different loss settings on CIFAR-10.}
\begin{center}
\begin{tabular}{c | c | c | c}
\hline\hline
Loss function & & Conv & PConv \\
\hline
\multirow{4}*{CE~\cite{goodfellow2014generative}} & trial 1& 16.72 & 14.10 \\
& trial 2 & 17.81 & 14.14 \\
& trial 3 & 16.65 & 13.80 \\
\cline{2-4}
& \textbf{Average} & $17.06 \pm 0.65$ & \textbf{14.02} $\pm$ \textbf{0.19}\\
\hline

\multirow{4}*{LSGAN~\cite{mao2017least}} & trial 1& 19.72 & 19.43 \\
& trial 2 & 19.60 & 18.82 \\
& trial 3 & 20.81 & 19.45 \\
\cline{2-4}
& \textbf{Average} & 20.04 $\pm$ 0.66 & \textbf{19.24 $\pm$ 0.36} \\

\hline
\multirow{4}*{Hinge~\cite{lim2017geometric}} & trial 1& 13.21 & 12.31 \\
& trial 2 & 13.44 & 12.58 \\
& trial 3 & 13.57 & 12.89 \\
\cline{2-4}
& \textbf{Average} & $13.41 \pm 0.18$ & \textbf{12.59} $\pm$ \textbf{0.29}\\

\hline\hline
\end{tabular}
\end{center}
\label{table:table5}
\end{table}

Indeed, as depicted in Fig.~\ref{fig:fig4}, SDrop produces discrete vectors since it turned on or off the all values in the selected dropping channels, \ie Bernoulli dropout. However, this problem could be mitigated by using the Gaussian dropout which perturbs the features by multiplying the scaling values sampled from $N\sim(1, p(1-p))$ distribution. This indicates that the SDrop with Gaussian dropout could cover the entire feature space like PConv. Thus, we measured the GAN performance when using SDrop with Gaussian dropout (we denoted it as $\textrm{SDrop}^\dagger$). As described in Table~\ref{table:table3}, $\textrm{SDrop}^\dagger$ is more stable than SDrop even though $\lambda_\textrm {R}$ becomes large. These results reveal that in order to train GAN stably, it is necessary to cover the entire feature space continuously. Although $\textrm{SDrop}^\dagger$ shows fine performance, the proposed method still outperforms the SDrop-based approaches. Based on these results, we concluded that PConv is more effective to boost the GAN performance compared to SDrop-based approaches. Since the performance of PConv is not significantly different when $\lambda_\textrm{R}=0.1$ and $\lambda_\textrm{R}=0.2$, in the rest of this paper, we conducted other experiments by setting $\lambda_\textrm{R}$ as 0.1.

\begin{table*}[!t]
\caption{Comparison of the proposed method with the standard convolutional layer on CIFAR-10, CelebA, LSUN, and tiny-ImageNet in terms of FID.}
\begin{center}
\begin{tabular}{c | c | c | c | c | c | c}
\hline\hline
\multirow{2}*{Dataset} & \multirow{2}*{Resolution} & & \multicolumn{2}{c|}{GAN} & \multicolumn{2}{c}{Conditional GAN}\\
\cline{4-7}
 &  &  & \multicolumn{1}{c|}{Conv} & \multicolumn{1}{c|}{PConv} & \multicolumn{1}{c|}{Conv} & \multicolumn{1}{c}{PConv}\\
\hline

\multirow{4}*{CIFAR-10} & \multirow{4}*{$32\times 32$} & trial 1 & 13.21 & 12.31 & 13.36 & 12.55 \\
 & &  trial 2 & 13.44 &  12.58 &  13.76 & 11.64 \\
& & trial 3 & 13.57 & 12.89 & 14.46 & 12.27 \\
\cline{3-7}
& & \textbf{Average} & 13.41 $\pm$ 0.18 & \textbf{12.59 $\pm$ 0.29} & 13.86 $\pm$ 0.56 & \textbf{12.15 $\pm$ 0.47} \\
\hline

\multirow{4}*{CelebA} & \multirow{4}*{$64\times 64$} & trial 1 & 6.10 & 5.50 & - & - \\
 & &  trial 2 & 5.90 &  5.44 &  - & - \\
& & trial 3 & 5.98 & 6.12 & - & - \\
\cline{3-7}
& & \textbf{Average} & 6.00 $\pm$ 0.10 & \textbf{5.69 $\pm$ 0.38} & - & - \\
\hline

\multirow{4}*{LSUN} & \multirow{4}*{$64\times 64$} & trial 1 & 19.10 & 16.58 & 17.32 & 16.91 \\
 & &  trial 2 & 19.40 &  15.38 & 17.42 & 16.73 \\
& & trial 3 & 19.50 & 16.84 & 18.78 & 16.24 \\
\cline{3-7}
& & \textbf{Average} & 19.33 $\pm$ 0.21 & \textbf{16.42 $\pm$ 0.52} & 17.87 $\pm$ 0.78 & \textbf{16.63 $\pm$ 0.34} \\
\hline

\multirow{4}*{tiny-ImageNet} & \multirow{4}*{$128\times 128$} & trial 1 & 55.44 & 51.46 & 34.53 & 34.06 \\
 & &  trial 2 & 60.66 &  50.90 & 34.98 & 32.93 \\
& & trial 3 & 59.68 & 48.90 & 34.59 & 33.61 \\
\cline{3-7}
& & \textbf{Average} & 58.59 $\pm$ 2.77 & \textbf{50.42} $\pm$ \textbf{1.34} & 34.70 $\pm$ 0.24 & \textbf{33.53} $\pm$ \textbf{0.57} \\
\hline

\hline\hline
\end{tabular}
\end{center}
\label{table:table6}
\end{table*}

Furthermore, we investigated whether PConv would alleviate the discriminator memorization problem. To verify whether the overfitting problem has occurred, the conventional image classification techniques generally measure the classification accuracy gap between the training and test sets. Therefore, a large gap in the classification accuracy is considered a more serious overfitting problem. Based on these approaches, we divided the CIFAR-10 dataset into the training and test sets, which contain 40,000 and 10,000 images, respectively. Thereafter, we trained a GAN using the training set as real samples. After completing the adversarial learning, we counted the number of images for which the discriminator outputs a value greater than zero because the discriminator was trained using the hinge-version of the adversarial loss~\cite{lim2017geometric}. As shown in Table~\ref{table:memorization}, the proposed method exhibits a lower accuracy gap between the training and test sets, compared with the standard convolutional layer. These observations indicate that PConv effectively alleviates the memorization problem in the discriminator. Although the SDrop moderates the memorization problem better than the proposed method, it is not effective for the adversarial learning as shown in Table~\ref{table:table3}. Thus, as mentioned in~\cite{brock2018large}, the conventional dropout techniques can effectively moderate the memorization problem; however, they often degrades the performance of GAN. Since the main goal is not only to alleviate the memorization problem but also boost the GAN performance, PConv is more suitable.

To further show the effectiveness of the proposed method, we trained the network using only the half or quarter number of images on the CIFAR-10 dataset. As shown in Table~\ref{table:table4}, the performance of the network using the standard convolutional layer is degraded when the number of training images becomes smaller. In contrast, the network trained with the proposed method achieves a better performance since PConv can moderate the memorization problem. Here, one may anticipate that the data augmentation (DA) techniques could alleviate the memorization problem. To clarify the superiority of PConv, we compared the performance of PConv against that of the standard convolution trained with DA. In our experiments, following the previous paper~\cite{zhao2020differentiable}, we built the augmented data using \textit{Translation} (with in [-1/8, 1/8] of the image size, padded with zero) and Cutout~\cite{devries2017improved} (making with a random square of half image size) techniques. As described in~\ref{table:table4}, the DA techniques improve the GAN performance, especially for training the GAN with a small number of real images, but they show weak performance compared with PConv. This means that the proposed method is suitable to train GAN using a small number of real images. 

In addition, to demonstrate the generalization ability of the proposed method, we trained the networks using various adversarial loss functions. In this paper, we conducted additional experiments using two different loss functions: the loss function based on the cross-entropy (CE) theorem (Eqs.~\ref{eq1:ganD} and~\ref{eq1:ganG}) and the loss function proposed in the least square GAN (LSGAN) paper~\cite{mao2017least}. As shown in Table~\ref{table:table5}, even with the various loss functions, the proposed method still has a superior performance, compared to the standard convolutional layer. These results reveal that the proposed method can be easily applied to the GAN without considering the experimental settings such as the adversarial loss function.

\begin{figure*}
\centering
\includegraphics[width=0.9\linewidth]{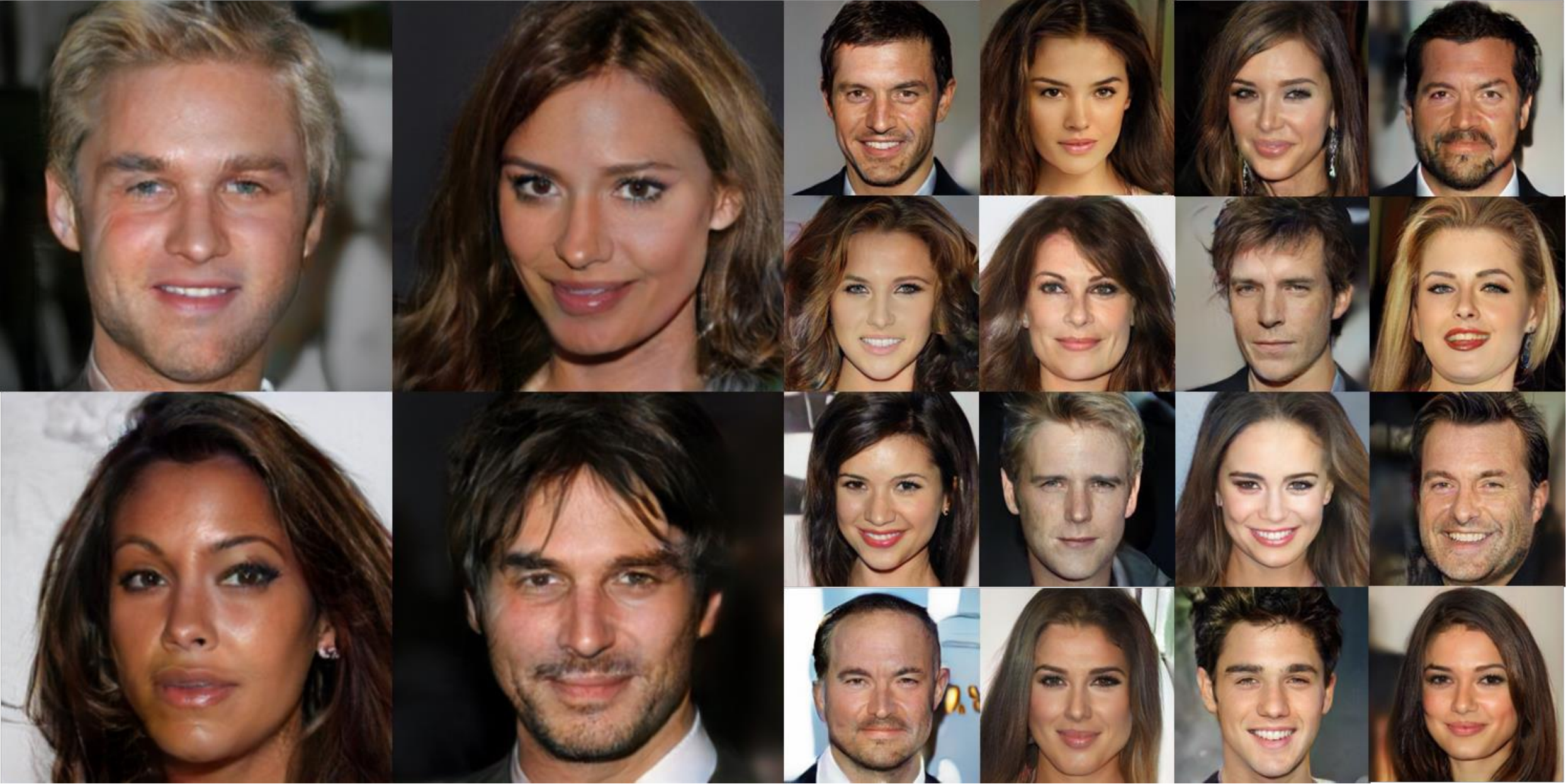}
\caption{Generated images with $512\times512$ and $256\times256$ resolutions on the Celeb-HQ dataset.}
\label{fig:fig5}
\vspace{-0cm}
\end{figure*}

Extensive experimental results on various datasets are summarized in Table~\ref{table:table6}. First, in the GAN scheme, the proposed method shows a superior performance than the standard convolutional layer. In particular, on the LSUN and tiny-ImageNet datasets containing complex images that are difficult to generate, the proposed method significantly improves the generator performance. These results indicate that even with spectral normalization, the discriminator using the standard convolution layer is struggle to learn robust features which are effective to guide the generator. By simply replacing the standard convolutional layer with the proposed method, we achieved a higher performance in the various datasets. The proposed method shows a slightly better performance than the standard convolutional layer on the CelebA dataset. Because low-resolution face images are easy to generate using the conventional techniques; it is difficult to further enhance the performance. However, when generating the high-resolution face images, the proposed method exhibits significantly superior performance than the conventional one. We will present the experimental results related to the high-resolution images later.

We conducted more experiments to validate the effectiveness of the proposed method in the conditional GAN scheme. Therefore, following the most representative conditional GAN scheme, we replaced the BN in the generator with the conditional BN layer~\cite{dumoulin2017learned} and added the conditional projection layer in the discriminator. Note that network architectures are the same as the models used for the experiments of GAN. As described in Table~\ref{table:table6}, similar to the trend of experimental results of GAN,  the proposed method exhibits a superior performance, compared to the conventional method~\cite{miyato2018cgans}. These results indicate that the proposed method can be applied to the conditional GAN scheme to boost the performance.

To show the effectiveness of the proposed method for generating high-resolution images, we conducted additional experiments using the CelebA-HQ dataset. In our experiments, the networks were trained to produce $256\times256$ and $512\times512$ images. The experimental results are presented in Table~\ref{table:table7} and Fig.~\ref{fig:fig5}. The proposed method shows significantly low FID scores, compared to the standard convolutional layer, and it produces visually pleasing images. These results demonstrate that the proposed method is also effective in generating high-resolution images. Indeed, this study does not intend to produce the design of an optimal generator and discriminator architectures for PConv. There can be another network architecture that improves the performance and generates more visually pleasing images. This paper focuses on verifying whether it is possible to achieve better performance by simply replacing the standard convolutions with PConv. 

\begin{table}[t]
\caption{Comparison of the proposed method and the standard convolutional layer on CelebA-HQ in terms of FID.}
\begin{center}
\begin{tabular}{c | c | c | c}
\hline\hline
Resolution & & Conv & PConv \\
\hline
\multirow{4}*{$256\times256$} & trial 1& 23.32 & 15.16 \\
& trial 2 & 21.98 & 14.08 \\
& trial 3 & 23.13 & 15.63 \\
\cline{2-4}
& \textbf{Average} & $22.81 \pm 0.72$ & \textbf{14.96} $\pm$ \textbf{0.80}\\
\hline

\multirow{4}*{$512\times512$} & trial 1& 28.11 & 18.26 \\
& trial 2 & 29.32 & 23.93 \\
& trial 3 & 28.18 & 20.34 \\
\cline{2-4}
& \textbf{Average} & $28.54 \pm 0.68$ & \textbf{20.84} $\pm$ \textbf{2.87} \\

\hline\hline
\end{tabular}
\end{center}
\label{table:table7}
\end{table}

\section{Conclusion and Future work}
\label{sec5}
This paper have introduced a straightforward technique for boosting the performance of GAN. By simply replacing the standard convolutional layer with PConv, the discriminator is able to effectively guide the generator, which results in performance improvement of the generator. The main advantage of the proposed method is that it can be easily applied to the existing discriminator networks without imposing the training overhead, while significantly improving the performance. Furthermore, this paper shows the generalization ability of PConv in various aspects through high-resolution image generation and several ablation studies. Therefore, we expect that PConv can be applicable to various GAN-based applications. 

Indeed, our manuscript focuses on introducing a novel approach specialized to the generative adversarial network (GAN). We agree that the perturbation procedure in PConv might work well for the discriminator, but might cause some issues in other networks used for different applications. Although our manuscript only covers the GAN, we have shown that the proposed method could improve the GAN performance significantly with various aspects. Thus, we expected that PConv could be effectively used for GAN-based diverse applications such as image-to-image translation. As our future work, we plan to further investigate a novel perturbation skill that covers various applications.

\section*{Declarations}
\textbf{Conflict of interest} The authors declare that they have no conflict of interest.

\bibliographystyle{spmpsci}      
\bibliography{egbib.bib}

\begin{thebibliography}{10}
\providecommand{\url}[1]{{#1}}
\providecommand{\urlprefix}{URL }
\expandafter\ifx\csname urlstyle\endcsname\relax
  \providecommand{\doi}[1]{DOI~\discretionary{}{}{}#1}\else
  \providecommand{\doi}{DOI~\discretionary{}{}{}\begingroup
  \urlstyle{rm}\Url}\fi

\bibitem{arjovsky2017wasserstein}
Arjovsky, M., Chintala, S., Bottou, L.: Wasserstein gan.
\newblock arXiv preprint arXiv:1701.07875  (2017)

\bibitem{brock2018large}
Brock, A., Donahue, J., Simonyan, K.: Large scale gan training for high
  fidelity natural image synthesis.
\newblock arXiv preprint arXiv:1809.11096  (2018)

\bibitem{choi2018stargan}
Choi, Y., Choi, M., Kim, M., Ha, J.W., Kim, S., Choo, J.: Stargan: Unified
  generative adversarial networks for multi-domain image-to-image translation.
\newblock In: Proceedings of the IEEE Conference on Computer Vision and Pattern
  Recognition, pp. 8789--8797 (2018)

\bibitem{23deng2009imagenet}
Deng, J., Dong, W., Socher, R., Li, L.J., Li, K., Fei-Fei, L.: Imagenet: A
  large-scale hierarchical image database.
\newblock In: 2009 IEEE conference on computer vision and pattern recognition,
  pp. 248--255. Ieee (2009)

\bibitem{devries2017improved}
DeVries, T., Taylor, G.W.: Improved regularization of convolutional neural
  networks with cutout.
\newblock arXiv preprint arXiv:1708.04552  (2017)

\bibitem{dumoulin2017learned}
Dumoulin, V., Shlens, J., Kudlur, M.: A learned representation for artistic
  style.
\newblock Proc. of ICLR \textbf{2} (2017)

\bibitem{goodfellow2014generative}
Goodfellow, I., Pouget-Abadie, J., Mirza, M., Xu, B., Warde-Farley, D., Ozair,
  S., Courville, A., Bengio, Y.: Generative adversarial nets.
\newblock In: Advances in neural information processing systems, pp. 2672--2680
  (2014)

\bibitem{gulrajani2017improved}
Gulrajani, I., Ahmed, F., Arjovsky, M., Dumoulin, V., Courville, A.C.: Improved
  training of wasserstein gans.
\newblock In: Advances in neural information processing systems, pp. 5767--5777
  (2017)

\bibitem{he2016deep}
He, K., Zhang, X., Ren, S., Sun, J.: Deep residual learning for image
  recognition.
\newblock In: Proceedings of the IEEE conference on computer vision and pattern
  recognition, pp. 770--778 (2016)

\bibitem{heusel2017gans}
Heusel, M., Ramsauer, H., Unterthiner, T., Nessler, B., Hochreiter, S.: Gans
  trained by a two time-scale update rule converge to a local nash equilibrium.
\newblock In: Advances in Neural Information Processing Systems, pp. 6626--6637
  (2017)

\bibitem{hong2018inferring}
Hong, S., Yang, D., Choi, J., Lee, H.: Inferring semantic layout for
  hierarchical text-to-image synthesis.
\newblock In: Proceedings of the IEEE Conference on Computer Vision and Pattern
  Recognition, pp. 7986--7994 (2018)

\bibitem{isola2017image}
Isola, P., Zhu, J.Y., Zhou, T., Efros, A.A.: Image-to-image translation with
  conditional adversarial networks.
\newblock In: Proceedings of the IEEE conference on computer vision and pattern
  recognition, pp. 1125--1134 (2017)

\bibitem{karras2017progressive}
Karras, T., Aila, T., Laine, S., Lehtinen, J.: Progressive growing of gans for
  improved quality, stability, and variation.
\newblock arXiv preprint arXiv:1710.10196  (2017)

\bibitem{karras2020training}
Karras, T., Aittala, M., Hellsten, J., Laine, S., Lehtinen, J., Aila, T.:
  Training generative adversarial networks with limited data.
\newblock Advances in Neural Information Processing Systems \textbf{33} (2020)

\bibitem{kingma2014adam}
Kingma, D.P., Ba, J.: Adam: A method for stochastic optimization.
\newblock arXiv preprint arXiv:1412.6980  (2014)

\bibitem{kodali2017convergence}
Kodali, N., Abernethy, J., Hays, J., Kira, Z.: On convergence and stability of
  gans.
\newblock arXiv preprint arXiv:1705.07215  (2017)

\bibitem{krizhevsky2012imagenet}
Krizhevsky, A., Sutskever, I., Hinton, G.E.: Imagenet classification with deep
  convolutional neural networks.
\newblock In: Advances in neural information processing systems, pp. 1097--1105
  (2012)

\bibitem{krizhevsky2017imagenet}
Krizhevsky, A., Sutskever, I., Hinton, G.E.: Imagenet classification with deep
  convolutional neural networks.
\newblock Communications of the ACM \textbf{60}(6), 84--90 (2017)

\bibitem{kurach2019large}
Kurach, K., Lu{\v{c}}i{\'c}, M., Zhai, X., Michalski, M., Gelly, S.: A
  large-scale study on regularization and normalization in gans.
\newblock In: International Conference on Machine Learning, pp. 3581--3590.
  PMLR (2019)

\bibitem{lim2017geometric}
Lim, J.H., Ye, J.C.: Geometric gan.
\newblock arXiv preprint arXiv:1705.02894  (2017)

\bibitem{liu2015deep}
Liu, Z., Luo, P., Wang, X., Tang, X.: Deep learning face attributes in the
  wild.
\newblock In: Proceedings of the IEEE International Conference on Computer
  Vision, pp. 3730--3738 (2015)

\bibitem{mao2017least}
Mao, X., Li, Q., Xie, H., Lau, R.Y., Wang, Z., Paul~Smolley, S.: Least squares
  generative adversarial networks.
\newblock In: Proceedings of the IEEE international conference on computer
  vision, pp. 2794--2802 (2017)

\bibitem{mescheder2018training}
Mescheder, L., Geiger, A., Nowozin, S.: Which training methods for gans do
  actually converge?
\newblock arXiv preprint arXiv:1801.04406  (2018)

\bibitem{mirza2014conditional}
Mirza, M., Osindero, S.: Conditional generative adversarial nets.
\newblock arXiv preprint arXiv:1411.1784  (2014)

\bibitem{miyato2018spectral}
Miyato, T., Kataoka, T., Koyama, M., Yoshida, Y.: Spectral normalization for
  generative adversarial networks.
\newblock arXiv preprint arXiv:1802.05957  (2018)

\bibitem{miyato2018cgans}
Miyato, T., Koyama, M.: cgans with projection discriminator.
\newblock arXiv preprint arXiv:1802.05637  (2018)

\bibitem{odena2017conditional}
Odena, A., Olah, C., Shlens, J.: Conditional image synthesis with auxiliary
  classifier gans.
\newblock In: Proceedings of the 34th International Conference on Machine
  Learning-Volume 70, pp. 2642--2651. JMLR. org (2017)

\bibitem{reed2016generative}
Reed, S., Akata, Z., Yan, X., Logeswaran, L., Schiele, B., Lee, H.: Generative
  adversarial text to image synthesis.
\newblock arXiv preprint arXiv:1605.05396  (2016)

\bibitem{roth2017stabilizing}
Roth, K., Lucchi, A., Nowozin, S., Hofmann, T.: Stabilizing training of
  generative adversarial networks through regularization.
\newblock In: Advances in neural information processing systems, pp. 2018--2028
  (2017)

\bibitem{sagong2019pepsi}
Sagong, M.c., Shin, Y.g., Kim, S.w., Park, S., Ko, S.j.: Pepsi: Fast image
  inpainting with parallel decoding network.
\newblock In: Proceedings of the IEEE Conference on Computer Vision and Pattern
  Recognition, pp. 11360--11368 (2019)

\bibitem{salimans2016improved}
Salimans, T., Goodfellow, I., Zaremba, W., Cheung, V., Radford, A., Chen, X.:
  Improved techniques for training gans.
\newblock In: Advances in neural information processing systems, pp. 2234--2242
  (2016)

\bibitem{shin2020pepsi++}
Shin, Y.G., Sagong, M.C., Yeo, Y.J., Kim, S.W., Ko, S.J.: Pepsi++: fast and
  lightweight network for image inpainting.
\newblock IEEE Transactions on Neural Networks and Learning Systems  (2020)

\bibitem{simard2003best}
Simard, P.Y., Steinkraus, D., Platt, J.C., et~al.: Best practices for
  convolutional neural networks applied to visual document analysis.
\newblock In: Icdar, vol.~3 (2003)

\bibitem{srivastava2014dropout}
Srivastava, N., Hinton, G., Krizhevsky, A., Sutskever, I., Salakhutdinov, R.:
  Dropout: a simple way to prevent neural networks from overfitting.
\newblock The journal of machine learning research \textbf{15}(1), 1929--1958
  (2014)

\bibitem{szegedy2016rethinking}
Szegedy, C., Vanhoucke, V., Ioffe, S., Shlens, J., Wojna, Z.: Rethinking the
  inception architecture for computer vision.
\newblock In: Proceedings of the IEEE conference on computer vision and pattern
  recognition, pp. 2818--2826 (2016)

\bibitem{tompson2015efficient}
Tompson, J., Goroshin, R., Jain, A., LeCun, Y., Bregler, C.: Efficient object
  localization using convolutional networks.
\newblock In: Proceedings of the IEEE conference on computer vision and pattern
  recognition, pp. 648--656 (2015)

\bibitem{27torralba200880}
Torralba, A., Fergus, R., Freeman, W.T.: 80 million tiny images: A large data
  set for nonparametric object and scene recognition.
\newblock IEEE transactions on pattern analysis and machine intelligence
  \textbf{30}(11), 1958--1970 (2008)

\bibitem{tran2020towards}
Tran, N.T., Tran, V.H., Nguyen, N.B., Nguyen, T.K., Cheung, N.M.: Towards good
  practices for data augmentation in gan training.
\newblock arXiv preprint arXiv:2006.05338  (2020)

\bibitem{wan2013regularization}
Wan, L., Zeiler, M., Zhang, S., Le~Cun, Y., Fergus, R.: Regularization of
  neural networks using dropconnect.
\newblock In: International conference on machine learning, pp. 1058--1066
  (2013)

\bibitem{yao2015tiny}
Yao, L., Miller, J.: Tiny imagenet classification with convolutional neural
  networks.
\newblock CS 231N  (2015)

\bibitem{yeo2021simple}
Yeo, Y.J., Shin, Y.G., Park, S., Ko, S.J.: Simple yet effective way for
  improving the performance of gan.
\newblock IEEE Transactions on Neural Networks and Learning Systems  (2021)

\bibitem{yu15lsun}
Yu, F., Zhang, Y., Song, S., Seff, A., Xiao, J.: Lsun: Construction of a
  large-scale image dataset using deep learning with humans in the loop.
\newblock arXiv preprint arXiv:1506.03365  (2015)

\bibitem{yu2018free}
Yu, J., Lin, Z., Yang, J., Shen, X., Lu, X., Huang, T.S.: Free-form image
  inpainting with gated convolution.
\newblock arXiv preprint arXiv:1806.03589  (2018)

\bibitem{zhang2017mixup}
Zhang, H., Cisse, M., Dauphin, Y.N., Lopez-Paz, D.: mixup: Beyond empirical
  risk minimization.
\newblock arXiv preprint arXiv:1710.09412  (2017)

\bibitem{zhang2018self}
Zhang, H., Goodfellow, I., Metaxas, D., Odena, A.: Self-attention generative
  adversarial networks.
\newblock arXiv preprint arXiv:1805.08318  (2018)

\bibitem{zhang2017stackgan}
Zhang, H., Xu, T., Li, H., Zhang, S., Wang, X., Huang, X., Metaxas, D.N.:
  Stackgan: Text to photo-realistic image synthesis with stacked generative
  adversarial networks.
\newblock In: Proceedings of the IEEE International Conference on Computer
  Vision, pp. 5907--5915 (2017)

\bibitem{zhang2018stackgan++}
Zhang, H., Xu, T., Li, H., Zhang, S., Wang, X., Huang, X., Metaxas, D.N.:
  Stackgan++: Realistic image synthesis with stacked generative adversarial
  networks.
\newblock IEEE transactions on pattern analysis and machine intelligence
  \textbf{41}(8), 1947--1962 (2018)

\bibitem{zhang2019consistency}
Zhang, H., Zhang, Z., Odena, A., Lee, H.: Consistency regularization for
  generative adversarial networks.
\newblock arXiv preprint arXiv:1910.12027  (2019)

\bibitem{zhao2020differentiable}
Zhao, S., Liu, Z., Lin, J., Zhu, J.Y., Han, S.: Differentiable augmentation for
  data-efficient gan training.
\newblock Advances in Neural Information Processing Systems \textbf{33} (2020)

\bibitem{zhao2020image}
Zhao, Z., Zhang, Z., Chen, T., Singh, S., Zhang, H.: Image augmentations for
  gan training.
\newblock arXiv preprint arXiv:2006.02595  (2020)

\bibitem{zhou2018don}
Zhou, B., Kr{\"a}henb{\"u}hl, P.: Don't let your discriminator be fooled.
\newblock In: International Conference on Learning Representations (2018)

\bibitem{zhu2017unpaired}
Zhu, J.Y., Park, T., Isola, P., Efros, A.A.: Unpaired image-to-image
  translation using cycle-consistent adversarial networks.
\newblock In: Proceedings of the IEEE international conference on computer
  vision, pp. 2223--2232 (2017)

\end{thebibliography}

\vspace{100mm}





\end{document}